\def\arxivversion{}
\documentclass[11pt,a4paper]{article}
\usepackage{times,latexsym}
\usepackage{url}
\makeatletter\g@addto@macro{\UrlBreaks}{\UrlOrds}\makeatother   %
\usepackage[T1]{fontenc}
\ifdefined\taclfinal\def\taclauthors{}\fi
\ifdefined\arxivversion\def\taclauthors{}\fi
\ifdefined\taclauthors\usepackage[acceptedWithA]{tacl2021v1}\else\usepackage{tacl2021v1}\fi
\ifdefined\taclauthors\fi
\hypersetup{pdftitle={Do We Need an LLM to Know Why Sources Conflict? Decision Models for Knowledge-Conflict Typing in RAG}}
\ifdefined\taclauthors\hypersetup{pdfauthor={Parv Khatri, Harshul Jain, Mehak Kapoor, Tanya Sah}}\fi
\newcommand{\repourl}{https://github.com/Parv17k/decision-models-conflict-typing}
\ifdefined\taclauthors\newcommand{\releasenote}{ at \url{\repourl}}\else\newcommand{\releasenote}{}\fi
\usepackage{xspace}
\xspaceaddexceptions{\% $ ] ) / , ;}   %
\usepackage{booktabs}
\usepackage{graphicx}
\usepackage{amsmath}
\usepackage{xcolor}
\usepackage{tikz}
\usepackage{listings}

\newsavebox{\fitbox}
\newcommand{\fitwidth}[1]{\sbox{\fitbox}{#1}\ifdim\wd\fitbox>\linewidth\typeout{FITWIDTH scaled \the\wd\fitbox\space to \the\linewidth}\resizebox{\linewidth}{!}{\usebox{\fitbox}}\else\usebox{\fitbox}\fi}
\makeatletter\let\dm@thebibliography\thebibliography
\renewcommand{\thebibliography}[1]{\dm@thebibliography{#1}\raggedright}\makeatother

\newcommand{\jev}{Jev\xspace}
\newcommand{\conflicts}{\textsc{Conflicts}\xspace}

\newcommand{\allgtwofive}{60.7\xspace}
\newcommand{\aurocmax}{0.76\xspace}
\newcommand{\aurocmin}{0.64\xspace}
\newcommand{\avgresults}{9.2\xspace}

\newcommand{\cascgainmax}{4.9\xspace}

\newcommand{\casclunatfacc}{67.0\xspace}

\newcommand{\casclunatfrand}{67.2\xspace}

\newcommand{\casctiedcostmax}{37\xspace}
\newcommand{\casctiedcostmin}{25\xspace}
\newcommand{\cihalf}{4.8\xspace}
\newcommand{\costratiorefmax}{39\xspace}
\newcommand{\costratiorefmin}{4\xspace}
\newcommand{\costratiorefnow}{2.2\xspace}
\newcommand{\devcompatcomp}{0.81\xspace}
\newcommand{\devcompatopin}{0.69\xspace}
\newcommand{\devmisinfohi}{0.79\xspace}
\newcommand{\devmisinfolo}{0.53\xspace}
\newcommand{\devnopin}{13\xspace}
\newcommand{\devopinionauroc}{0.83\xspace}
\newcommand{\devopintocomp}{12\xspace}

\newcommand{\diaggmrerunflip}{3.4\xspace}
\newcommand{\diaggmshuffleflip}{9.4\xspace}

\newcommand{\diaglunashuffleonediff}{$-$4.7\xspace}

\newcommand{\diaglunashuffletwodiff}{$-$5.7\xspace}

\newcommand{\dmaccmax}{67.0\xspace}
\newcommand{\dmaccmedian}{62.6\xspace}
\newcommand{\dmaccmin}{50.0\xspace}
\newcommand{\dmcostmax}{0.41\xspace}
\newcommand{\dmcostmin}{0.04\xspace}
\newcommand{\dmlatmax}{0.72\xspace}
\newcommand{\dmlatmin}{0.26\xspace}
\newcommand{\ecemaximum}{0.188\xspace}
\newcommand{\eceminimum}{0.031\xspace}
\newcommand{\eecascdiffmax}{+0.7\xspace}

\newcommand{\eecascdiffmin}{$-$1.0\xspace}

\newcommand{\eedmdiffmax}{+0.7\xspace}

\newcommand{\eedmdiffmin}{$-$3.9\xspace}

\newcommand{\eenanswers}{2,436\xspace}
\newcommand{\eencasctied}{11\xspace}
\newcommand{\eenrecall}{202\xspace}
\newcommand{\eentied}{10\xspace}

\newcommand{\eeoracle}{95.3\xspace}

\newcommand{\eeoracleopin}{97.1\xspace}
\newcommand{\eeoracleoutd}{87.3\xspace}

\newcommand{\eeoraclevsvandiff}{+7.4\xspace}

\newcommand{\eeoraclevsvanp}{<0.001\xspace}
\newcommand{\eepipe}{91.6\xspace}

\newcommand{\eepipevsvandiff}{+3.7\xspace}

\newcommand{\eepipevsvanp}{=0.032\xspace}

\newcommand{\eerecallmax}{93.1\xspace}
\newcommand{\eerecallmin}{90.6\xspace}

\newcommand{\eescaledownvspipediff}{$-$3.9\xspace}

\newcommand{\eescaledownvspipep}{=0.011\xspace}

\newcommand{\eevalacc}{92\xspace}
\newcommand{\eevalaccone}{91\xspace}
\newcommand{\eevalacctwo}{87\xspace}
\newcommand{\eevalkappa}{0.46\xspace}
\newcommand{\eevan}{87.9\xspace}

\newcommand{\eevanopin}{82.4\xspace}
\newcommand{\eevanoutd}{72.7\xspace}

\newcommand{\fbntied}{11\xspace}

\newcommand{\fboraclevsvandiff}{+5.7\xspace}

\newcommand{\fboraclevsvanp}{<0.001\xspace}
\newcommand{\fbpipe}{89.7\xspace}

\newcommand{\fbpipevsvanp}{=0.47\xspace}

\newcommand{\fbvalacc}{87\xspace}
\newcommand{\fbvalaccone}{84\xspace}
\newcommand{\fbvalacctwo}{82\xspace}

\newcommand{\fbvan}{91.1\xspace}

\newcommand{\gthinkvsnothinkdiff}{$-$3.2\xspace}

\newcommand{\gthinkvsnothinkp}{=0.019\xspace}
\newcommand{\gthreeeightacc}{70.0\xspace}

\newcommand{\gthreeeightdefaultacc}{66.7\xspace}

\newcommand{\gthreeeightmisif}{25.0\xspace}
\newcommand{\gthreeeightoutdf}{66.1\xspace}
\newcommand{\gthreeeightvstwofivediff}{+8.6\xspace}

\newcommand{\gthreeeightvstwofivep}{<0.001\xspace}
\newcommand{\gtwofiveacc}{61.3\xspace}
\newcommand{\gtwofivecost}{1.66\xspace}

\newcommand{\gtwofivelat}{1.55\xspace}

\newcommand{\gtwofivethinkacc}{57.9\xspace}

\newcommand{\gtwofivethinkvsnothinkp}{=0.19\xspace}
\newcommand{\humanfollows}{90\xspace}
\newcommand{\iaa}{92\xspace}
\newcommand{\iaakappa}{0.67\xspace}

\newcommand{\jevece}{0.103\xspace}
\newcommand{\jevflatacc}{61.8\xspace}

\newcommand{\jevtreeacc}{63.5\xspace}
\newcommand{\jevtreeece}{0.026\xspace}

\newcommand{\jevtreevsflatp}{=0.59\xspace}

\newcommand{\latratiorefmax}{6.0\xspace}
\newcommand{\latratiorefmin}{2.6\xspace}

\newcommand{\liquidultravsnowdiffabs}{3.0\xspace}

\newcommand{\liquidultravsrefdiff}{+5.7\xspace}

\newcommand{\liquidultravsrefp}{=0.027\xspace}

\newcommand{\lunaauroc}{0.65\xspace}

\newcommand{\lunavsnowdiffabs}{3.7\xspace}

\newcommand{\misifmax}{11.8\xspace}

\newcommand{\nall}{458\xspace}
\newcommand{\natpar}{10\xspace}
\newcommand{\nbeatrandom}{10\xspace}
\newcommand{\nbetterref}{2\xspace}
\newcommand{\ncasctied}{8\xspace}

\newcommand{\ndatesig}{0\xspace}
\newcommand{\ndatetests}{22\xspace}
\newcommand{\ndeterministic}{5\xspace}
\newcommand{\ndev}{52\xspace}
\newcommand{\ndm}{11\xspace}

\newcommand{\nflipless}{1\xspace}
\newcommand{\nflipmore}{4\xspace}
\newcommand{\nflipsame}{6\xspace}
\newcommand{\nfree}{2\xspace}
\newcommand{\ngemini}{14,470\xspace}
\newcommand{\nlenient}{10527\xspace}
\newcommand{\nmisitest}{4\xspace}
\newcommand{\noverconf}{7\xspace}
\newcommand{\nplant}{96\xspace}
\newcommand{\nplantaurochigh}{10\xspace}
\newcommand{\nproviders}{7\xspace}

\newcommand{\nshufflesig}{2\xspace}
\newcommand{\ntest}{406\xspace}
\newcommand{\ntiednow}{3\xspace}
\newcommand{\ntiedref}{8\xspace}
\newcommand{\nvalid}{100\xspace}
\newcommand{\nvoneworse}{11\xspace}

\newcommand{\nvtwotied}{7\xspace}
\newcommand{\nvtwoworse}{4\xspace}
\newcommand{\nworsenow}{8\xspace}
\newcommand{\nworseref}{1\xspace}

\newcommand{\nyoudenlower}{8\xspace}
\newcommand{\nyoudentied}{3\xspace}
\newcommand{\outdfmax}{72.7\xspace}
\newcommand{\outdfmin}{27.5\xspace}
\newcommand{\plantaurocmax}{0.86\xspace}
\newcommand{\plantaurocmin}{0.66\xspace}

\newcommand{\plantgmhit}{52.1\xspace}
\newcommand{\plantgmyouden}{+0.490\xspace}

\newcommand{\planthitmax}{66.7\xspace}
\newcommand{\planthitmin}{7.3\xspace}

\newcommand{\pplxvsrefdiff}{+5.4\xspace}

\newcommand{\pplxvsrefp}{=0.049\xspace}
\newcommand{\rerunflipmax}{9.4\xspace}

\newcommand{\scaledownvsrefdiff}{$-$11.3\xspace}

\newcommand{\scaledownvsrefp}{<0.001\xspace}
\newcommand{\shuffleflipmax}{42.1\xspace}
\newcommand{\shuffleflipmin}{5.4\xspace}
\newcommand{\testnopin}{102\xspace}
\newcommand{\testopintocomp}{89\xspace}
\newcommand{\treeoneopinf}{3.8\xspace}
\newcommand{\versionsdiff}{$-$2.2\xspace}

\newcommand{\versionsp}{=0.12\xspace}
\newcommand{\versionstreediff}{$-$0.7\xspace}

\newcommand{\versionstreep}{=0.66\xspace}
\newcommand{\youdenmax}{+0.490\xspace}

\newcommand{\youdenmin}{+0.052\xspace}

\title{Do We Need an LLM to Know Why Sources Conflict?\\
Decision Models for Knowledge-Conflict Typing in RAG}

\newcommand{\authoremail}[1]{{\small\texttt{#1}}}
\author{
  Parv Khatri \\ Independent Researcher \\ \authoremail{khatriparv@gmail.com}
  \And
  Harshul Jain \\ Independent Researcher \\ \authoremail{harshuljain1393@gmail.com}
  \AND
  Mehak Kapoor \\ Independent Researcher \\ \authoremail{mehakkapoor9873@gmail.com}
  \And
  Tanya Sah \\ Independent Researcher \\ \authoremail{tanyasah20@gmail.com}
}

\date{}

\begin{document}
\maketitle

\begin{abstract}
When the sources retrieved for a query disagree, the right response depends on \emph{why} they disagree: a
question with several valid answers, a contested topic, stale information and a false source each call for
different behaviour. \citet{cattan2025dragged} showed that telling a generator the conflict type improves its
responses by about 24 points, but that predicting the type is hard even for frontier LLMs. We study whether
\emph{typed decision models}, which return probabilities over fixed options in a single forward pass instead of
generating text, can do this classification fast and cheaply. Using the \conflicts benchmark and the original
paper's prompt verbatim, we evaluate all \ndm decision models served by one commercial inference gateway, from
\nproviders providers, against Gemini models on identical inputs. Asked directly, the decision models reach
\dmaccmin--\dmaccmax\% accuracy; \natpar of the \ndm match or exceed Gemini~2.5 Flash, the paper's model
(\gtwofiveacc\%), at \costratiorefmin--\costratiorefmax$\times$ lower billed cost (\nfree were free) and
\latratiorefmin--\latratiorefmax$\times$ lower latency. The current Gemini~3.8 Flash (\gthreeeightacc\%) is significantly more accurate than \nworsenow of
them, but sending each decision model's least confident quarter of queries to it gives a cascade statistically tied
with it for \ncasctied of the \ndm, at \casctiedcostmin--\casctiedcostmax\% of its cost. End to end, answers
generated with the types of \eentied of the \ndm follow the expected behaviour as often as with the LLM's types. Decomposing the taxonomy into yes/no questions lowers
accuracy for every model, and a repair fitted on one model does not transfer. Pre-specified diagnostics show that
dates play no measurable role and that sensitivity to source order varies widely between models. Decision models
detect a planted false source less reliably than the LLM as hard labels, but their probabilities still separate
planted from clean sources. We release code, prompts and
all model outputs\releasenote.

\end{abstract}

\section{Introduction}
\label{sec:intro}

Search-augmented language models answer from retrieved web pages, and those pages often disagree. The
disagreement can have very different causes. A question may be underspecified and admit several correct answers;
it may touch a contested topic on which reasonable sources differ; the answer may have changed over time, so some
pages are stale; or a page may simply be wrong. Each case calls for a different response: combine the answers,
present the viewpoints neutrally, prefer the most recent information, or ignore the unreliable source.
\citet{cattan2025dragged} formalised this as a five-way taxonomy, released \conflicts, a benchmark of 458 search
queries annotated by experts with their conflict type, and showed that giving a generator the gold type raises
the share of responses with the expected behaviour by about 24 points. Predicting the type is harder: the best LLM
they tested, Gemini~2.5 Flash, reached 65.3\% accuracy.

Conflict typing is a classification problem. It is made once per query, before the answer is written, and its
output is one of five labels. A classifier for it should therefore be fast enough to sit in front of every
search-augmented request, cheap enough to run on all of them, and calibrated enough that its uncertainty can be
acted on. A recent class of models, \emph{typed decision models}, is designed for this setting: they take
unstructured context and typed questions and return probabilities over the allowed answers in a single forward
pass, without generating text \citep{typesafe2026jev}. Since the first such model, several
providers have released their own, and one inference gateway now serves \ndm of them behind a common interface. Existing
evaluations cover classification, NLI and judging benchmarks, mostly for a single model
\citep{deusser2026jev,li2026specialized,duan2026general}, but none covers deciding why retrieved sources conflict,
a task that requires reading many documents at once and reasoning about their mutual relation rather than about
one text.

\begin{figure}[t]
\centering
\begin{tikzpicture}[
  font=\scriptsize, >=stealth, line width=0.6pt,
  box/.style={draw=black!65, rounded corners=2pt, align=center, inner sep=2pt, minimum height=10mm},
  note/.style={font=\scriptsize\itshape, text=black!60, align=center, inner sep=1pt},
  arr/.style={->, draw=black!75}]
  \node[box, fill=black!5, minimum width=14mm] (in) at (0, 0) {Query and\\$\sim$10 search\\results};
  \node[box, fill=blue!12, minimum width=23mm] (dm) at (2.75, 0) {\textbf{Decision model}\\one forward pass,\\$P(\text{type})$ over 5};
  \node[box, fill=orange!20, minimum width=23mm] (llm) at (2.75, -1.8) {\textbf{LLM}\\original prompt,\\generated text};
  \node[box, fill=black!5, minimum width=11mm] (type) at (5.75, -0.9) {Conflict\\type};
  \draw[arr] (in) -- (dm);
  \draw[arr] (dm.east) -- node[note, above, sloped] {confident} (type.north west);
  \draw[arr] (dm.south) -- node[note, right, align=left] {least confident\\$f$ (cascade)} (llm.north);
  \draw[arr] (llm.east) -- (type.south west);
  \draw[arr, dashed, draw=black!45] (in.south) |- (llm.west);
  \node[box, fill=black!5, minimum width=27mm] (gen) at (4.85, -3.85) {\textbf{Generator}\\answers from the\\results, given the type};
  \node[box, fill=black!5, minimum width=27mm] (judge) at (1.0, -3.85) {\textbf{Judge}\\does the answer show\\the expected behaviour?};
  \draw[arr] (type.south) -- (type.south |- gen.north);
  \draw[arr] (gen) -- (judge);
  \draw[black!30, densely dotted] (-0.8, -2.7) -- (6.45, -2.7);
  \node[note, anchor=north west] at (-0.8, 1.0) {Classification: \S\ref{sec:main}--\S\ref{sec:diagnostics}};
  \node[note, anchor=north west] at (-0.8, -2.75) {End to end: \S\ref{sec:endtoend}};
\end{tikzpicture}
\caption{Overview. A decision model types the conflict among a query's search results in one forward pass; in a
cascade, its least confident fraction $f$ of queries is typed by an LLM instead. End to end, a generator answers
given the predicted type, and a judge checks the answer against the expected behaviour of the gold type.}
\label{fig:overview}
\end{figure}
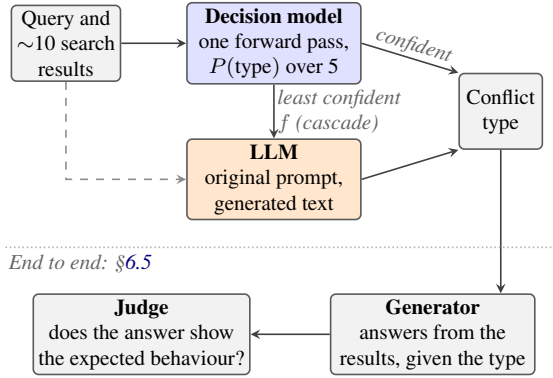

Can decision models, as a class, serve as this decision layer? And if only some can, what sets them apart? Our
study (Figure~\ref{fig:overview}) makes five contributions.

\paragraph{(1) A census of decision models on conflict typing.} We evaluate every decision model served by one
commercial gateway (\ndm models from \nproviders providers) on \conflicts under the original paper's protocol: the
same input representation, the same category definitions and examples taken verbatim from the paper's source, and
Gemini rerun on identical inputs so that every comparison is paired (\S\ref{sec:main}). \natpar of the \ndm
models match or exceed the paper's LLM at a fraction of its cost and latency; the current Gemini~3.8 Flash is
significantly more accurate than \nworsenow of them.

\paragraph{(2) A formulation study.} Decomposing the taxonomy into yes/no questions lowers accuracy for all \ndm
models, and a repair fitted on one model helps none, leaving the direct question as the reliable format
(\S\ref{sec:formulation}).

\paragraph{(3) Cost-quality cascades.} Decision models' confidence ranks their own errors, so sending their least
certain queries to an LLM is effective: with a quarter of the queries escalated to Gemini~3.8 Flash, \ncasctied of
the \ndm cascades are statistically tied with the LLM alone at \casctiedcostmin--\casctiedcostmax\% of its cost
(\S\ref{sec:cascade}).

\paragraph{(4) Pre-specified diagnostics.} We test whether decision models depend on publication dates,
are sensitive to source order, or miss a single false source among agreeing ones (\S\ref{sec:diagnostics}). Dates
play no measurable role, order sensitivity varies from model to model, and a planted false source shows in the
probabilities more than in the top answer.

\paragraph{(5) An end-to-end test.} Given to a generator, the types of \eentied of the \ndm decision models yield
answers that follow the expected behaviour as often as with the LLM's types, under both judges we specified
(\S\ref{sec:endtoend}).

Every setting and every later addition was fixed in writing before it was run (\S\ref{sec:setup}). We release the
code, prompts, splits and all model outputs\releasenote.

\section{Related Work}
\label{sec:related}

\paragraph{Knowledge conflicts in retrieval-augmented generation.}
\citet{xu2024knowledge} survey conflicts between a model's parametric knowledge and its context, among retrieved
contexts, and within the model's memory. Work on inter-context conflicts has largely studied one cause at a time,
such as substituted entities \citep{longpre2021entity}, contradictions in Wikipedia \citep{hou2024wikicontradict},
contested questions \citep{wan2024evidence}, temporal or geographic context \citep{zhang2021situatedqa}, freshness
\citep{vu2023freshllms} or conflicting contexts in open-domain QA \citep{liu2024qacc}. \citet{cattan2025dragged}
unify these causes into a taxonomy with expected behaviours and release \conflicts, the first benchmark annotated
with the conflict type; we build directly on it. Methods for resolving conflicts include multi-agent debate over
documents \citep{wang2025madam}, reasoning traces learned by fine-tuning \citep{mishra2025facts}, and a two-stage
detector that sends an embedding classifier's uncertain cases to an LLM \citep{wang2026conflictrag}. None of these
uses a typed decision model.

\paragraph{Typed decision models.}
A typed decision model answers typed questions (choice, score, yes/no) over a supplied state with probabilities
over the allowed answers, trained for calibration rather than generation. The class was introduced with \jev
\citep{typesafe2026jev}, and hosted and open-weight alternatives followed; open reimplementations have been
benchmarked against the hosted reference \citep{weng2026ownjev}. Independent evaluations, so far mostly of the
first model, find decision models competitive with frontier LLMs on knowledge, classification and NLI benchmarks at
a fraction of the cost \citep{deusser2026jev,li2026specialized}, calibrated on choice questions but biased towards
\emph{yes} when many yes/no questions share a request \citep{deusser2026jev,li2026llm2jev}, not always coherent
across logically related questions \citep{li2026beyond}, weaker as inputs grow long \citep{duan2026general}, and
vulnerable to natural-looking context that flips their decisions \citep{xu2026jevout}. As judges, they are cheaper
than LLMs but share their most confident errors \citep{rao2026rubric}, and a confidence-gated cascade to an LLM can
match the LLM at lower cost \citep{li2026jevjudge}. Closest to our setting, \citet{huang2026radiology} use a decision
model to decide whether one report supports or contradicts another. \citet{duan2026general} compare several decision
models on general benchmarks; no prior work compares them on typing the conflict among many retrieved sources.

\paragraph{Cascades and selective prediction.}
Routing easy inputs to a cheap model and hard ones to an expensive model is an established way to trade cost for
quality \citep{chen2023frugalgpt}; its effectiveness depends on how well the cheap model's confidence ranks its own
errors, the quantity studied in selective classification \citep{geifman2017selective}. We measure this directly
for every decision model on conflict typing and compare against random escalation.

\section{Task and Data}
\label{sec:task}

\paragraph{Taxonomy.} We use the taxonomy of \citet{cattan2025dragged}. Given a query and its retrieved sources,
the conflict is one of: \emph{No conflict} (the sources agree, up to wording or granularity);
\emph{Complementary information} (the answers differ but can all be correct, e.g.\ because the query is
underspecified); \emph{Conflicting opinions or research outcomes} (a subjective or contested topic with genuinely
opposing views); \emph{Outdated information} (a factual question on which sources disagree because the answer
changed over time); and \emph{Misinformation} (a factual question on which some sources are wrong).

\paragraph{Data.} \conflicts contains \nall queries drawn from ConflictingQA
\citep{wan2024evidence}, the geographic and temporal subsets of SituatedQA \citep{zhang2021situatedqa}, FreshQA
\citep{vu2023freshllms} and QACC \citep{liu2024qacc}, each with the top Google Search results
(\avgresults on average), the conflict type assigned by expert annotators, and, for the types with a single correct
answer, that answer. Each result carries a URL, title, snippet, publication date when available, and the
512-token segment of the page most relevant to the query. The released file has no split. We draw a stratified
development split of \ndev queries with a fixed seed and use it only to choose settings; the remaining \ntest
queries form the test split on which we report (Table~\ref{tab:data}). \emph{Misinformation} has only five
instances in the whole dataset, a limitation we return to in \S\ref{sec:diagnostics} by constructing a planted
variant.

\begin{table}[t]
\centering\small
\fitwidth{\begin{tabular}{lrrr}
\toprule
Conflict type & Dev & Test & All \\
\midrule
No conflict & 18 & 143 & 161 \\
Complementary information & 13 & 102 & 115 \\
Conflicting opinions & 13 & 102 & 115 \\
Outdated information & 7 & 55 & 62 \\
Misinformation & 1 & 4 & 5 \\
\midrule
Total & 52 & 406 & 458 \\
\bottomrule
\end{tabular}}
\caption{\conflicts \citep{cattan2025dragged} and our frozen split. The development split (stratified, fixed seed) is used only to choose settings; all reported results are on the test split unless stated.}
\label{tab:data}
\end{table}

\paragraph{Metric.} Following \citet{cattan2025dragged}, the primary metric is five-class accuracy. We also report
macro-F1 (the classes are imbalanced), per-class F1, and, for systems that return probabilities, expected
calibration error \citep[ECE;][]{guo2017calibration} of the top-class probability.

\section{Conflict Typing with Decision Models}
\label{sec:method}

A typed decision model receives a \emph{state}, here the query and its numbered search results, and one or more
typed questions. For a \emph{choice} question it returns a probability for each option; for a \emph{yes/no}
question, the probability of \emph{yes}. Several questions over the same state are answered in one request. We
study two ways of asking for the conflict type.

\paragraph{Flat.} One choice question whose five options are the five categories. The instruction, the category
definitions and their examples are copied word for word from the prompt of \citet{cattan2025dragged}
(Appendix~\ref{app:prompts}); a test in our code fails if a single word differs. The predicted type is the
most probable option, and its probability is the model's confidence.

\paragraph{Tree.} The taxonomy is hierarchical, which suggests asking simpler yes/no questions in one request:
\emph{differ} (do the results give different answers?), \emph{compatible} (could all of them be correct?),
\emph{opinion} (is it a contested topic with opposing views?), \emph{outdated} (do they disagree because the answer
changed?), and \emph{misinformation} (are some results incorrect?). The five probabilities are combined in code into
a distribution over the categories. \textbf{Tree v1} follows the taxonomy's order: $P(\textit{No conflict}) =
1-d$, $P(\textit{Complementary}) = d\,c$, and the remaining mass $d(1-c)$ is split over the three causes in
proportion to their yes-probabilities, where $d$ and $c$ are the probabilities of \emph{differ} and
\emph{compatible}. \textbf{Tree v2}, chosen after inspecting \jev~1.13's v1 answers on the development split (\S\ref{sec:formulation}),
asks \emph{opinion} before \emph{compatible} and treats \emph{misinformation} as the factual conflict not explained
by time: $P(\textit{Opinions}) = d\,o$, $P(\textit{Complementary}) = d(1-o)c$, $P(\textit{Outdated}) =
d(1-o)(1-c)\,t$ and $P(\textit{Misinformation}) = d(1-o)(1-c)(1-t)$, with $o$ and $t$ the probabilities of
\emph{opinion} and \emph{outdated}. Both trees use the same five answers, so v2 costs nothing extra. Every decision model receives exactly the same
questions and combination rules; nothing is tuned per model.

\paragraph{Cascade.} The decision model answers every query. The fraction $f$ of queries with its lowest confidence
is sent to an LLM, whose answer replaces the decision model's. The threshold is a quantile of the decision model's
confidences and uses no labels, so the
whole accuracy--cost curve can be computed on the test split without leakage; we fixed $f=25\%$ and $f=50\%$ as the
reported operating points before computing any cascade.

\section{Experimental Setup}
\label{sec:setup}

\paragraph{Inputs.} Every system sees the same input. As in \citet{cattan2025dragged}, each search result is the
concatenation of its URL, title, snippet, publication date (when available) and 512-token segment; results keep the
dataset's order and are numbered. For the decision models they form the state; for the LLMs they fill the paper's
prompt.

\paragraph{Decision models.} We run every model that the Vercel AI Gateway\footnote{\url{https://vercel.com/ai-gateway}} listed as a
decision model on the day of the census (25 October 2026),
with a pinned identifier where one exists: \jev~1.13.0 and 1.14.0 \citep{typesafe2026jev}, Decider v1.2 and Decider Flash v1.2
\citep{perplexity2026decider,perplexity2026deciderflash}, GPT-6 Luna Decisions \citep{openai2026decisions}, d1 and d1
Ultra \citep{liquid2026d1,liquid2026d1ultra}, Decision-1 \citep{microsoft2026decision}, Elephant-1 and Mammoth-1
\citep{baseten2026decision}, and Classify-1 \citep{scaledown2026classify}. All receive the questions of \S\ref{sec:method} unchanged.
Classify-1's API accepts only a single text field, so it receives the same content as one text, the numbered
results exactly as the LLMs see them. Two entries were excluded before any run. Laya, the only open decision model
served by the gateway, cannot accept the task's question, whose category definitions exceed its input limit; prior
work reports it near chance without task-specific training \citep{duan2026general,nandakishore2026laya}.
\texttt{typesafe-ai/jev} is a floating alias of a pinned version we already run. Open-weight models that the gateway
does not serve are out of scope: their cost and latency would depend on local hardware and not be comparable.

\paragraph{LLM baselines.} We rerun the paper's best classifier, Gemini~2.5 Flash \citep{comanici2025gemini}, with the paper's prompt
verbatim, temperature 0, thinking off (the paper lists Gemini 2.5 Flash and its thinking variant separately) and,
as a sensitivity check, thinking on. To compare against a current model we add Gemini~3.8 Flash, the newest Flash
model available to us, at its lowest thinking level (it cannot be switched off) and at its default. The prompt asks
for JSON with an explanation and a category number; answers stating the category in another format (\nlenient of
\ngemini) are parsed leniently rather than counted as errors. All models are reached through this gateway, which reports the billed cost of every request (prices may
differ from the providers' own APIs); latency is the wall-clock time of the successful request.

\paragraph{Protocol.} Prompts, questions, inputs and the development/test split were frozen before any test-set
call, with file hashes recorded. The only setting chosen by inspecting data is tree v2, chosen on the development
split and marked $\dagger$ in all tables. Every later addition (Gemini~3.8 Flash, the further decision models, the
cascade, the diagnostics and the end-to-end test) was specified in writing before it was run, and every run is reported
(Appendix~\ref{app:protocol}).

\paragraph{Statistics.} Comparisons are paired on identical queries. We report 95\% bootstrap intervals over queries
\citep{efron1994bootstrap} (10,000 resamples) and two-sided exact McNemar tests \citep{mcnemar1947} on per-query
correctness. With \ntest test queries, a 95\% interval on accuracy is about $\pm$\cihalf points wide, so differences
under about five points are generally not significant; we report them as ties. We do not correct for multiple
comparisons; with \ndm models a few nominal differences at $p<0.05$ are expected by chance, so we draw conclusions
from patterns across models rather than from single tests.

\section{Results}
\label{sec:results}

\begin{table*}[t]
\centering\small\setlength{\tabcolsep}{4.5pt}
\fitwidth{\begin{tabular}{lccccccc}
\toprule
System & Accuracy [95\% CI] & Macro-F1 & ECE & \$ / 1k & Lat.\ (s) & $\Delta$ Gem.\ 2.5 ($p$) & $\Delta$ Gem.\ 3.8 ($p$) \\
\midrule
\multicolumn{8}{l}{\emph{Decision models, flat question (one forward pass)}} \\
\quad Liquid d1 Ultra & 67.0 [62.3, 71.4] & 53.5 & 0.044 & 0.18 & 0.46 & +5.7 (0.027) & $-$3.0 (0.22) \\
\quad Perplexity Decider v1.2 & 66.7 [62.1, 71.2] & 56.5 & 0.061 & 0.09 & 0.42 & +5.4 (0.049) & $-$3.2 (0.14) \\
\quad OpenAI GPT-6 Luna Decisions & 66.3 [61.6, 70.9] & 54.7 & 0.059 & 0.41 & 0.35 & +4.9 (0.090) & $-$3.7 (0.13) \\
\quad Baseten Mammoth-1 & 65.8 [61.1, 70.4] & 52.9 & 0.052 & 0$^\P$ & 0.59 & +4.4 (0.10) & $-$4.2 (0.050) \\
\quad Baseten Elephant-1 & 65.3 [60.8, 69.7] & 51.6 & 0.040 & 0$^\P$ & 0.48 & +3.9 (0.16) & $-$4.7 (0.040) \\
\quad Liquid d1 & 62.6 [57.6, 67.2] & 51.5 & 0.075 & 0.08 & 0.41 & +1.2 (0.71) & $-$7.4 (0.002) \\
\quad Jev 1.13 & 61.8 [56.9, 66.5] & 54.9 & 0.103 & 0.19 & 0.26 & +0.5 (0.92) & $-$8.1 ({<}0.001) \\
\quad Perplexity Decider Flash v1.2 & 61.8 [57.1, 66.5] & 49.4 & 0.031 & 0.04 & 0.31 & +0.5 (0.93) & $-$8.1 ({<}0.001) \\
\quad Jev 1.14 & 59.6 [54.7, 64.5] & 53.3 & 0.113 & 0.19 & 0.28 & $-$1.7 (0.56) & $-$10.3 ({<}0.001) \\
\quad Microsoft Decision-1 & 58.1 [53.2, 62.8] & 50.2 & 0.188 & 0.18 & 0.45 & $-$3.2 (0.24) & $-$11.8 ({<}0.001) \\
\quad ScaleDown Classify-1$^\S$ & 50.0 [45.1, 54.9] & 37.4 & 0.090 & 0.24 & 0.72 & $-$11.3 ({<}0.001) & $-$20.0 ({<}0.001) \\
\multicolumn{8}{l}{\emph{Generative LLMs, the paper's prompt}} \\
\quad Gemini 2.5 Flash, thinking off & 61.3 [56.7, 66.0] & 53.1 & -- & 1.66 & 1.55 & -- & $-$8.6 ({<}0.001) \\
\quad Gemini 2.5 Flash, thinking on & 57.9 [53.0, 62.8] & 51.4 & -- & 6.04 & 9.57 & $-$3.4 (0.19) & $-$12.1 ({<}0.001) \\
\quad Gemini 3.8 Flash, thinking low & 70.0 [65.5, 74.4] & 61.3 & -- & 3.65 & 2.18 & +8.6 ({<}0.001) & -- \\
\quad Gemini 3.8 Flash, default thinking & 66.7 [62.1, 71.2] & 59.4 & -- & 5.51 & 5.01 & +5.4 (0.043) & $-$3.2 (0.019) \\
\midrule
\quad Gemini 2.5 Flash, as published$^\ddagger$ & 65.3 & -- & -- & -- & -- & -- & -- \\
\bottomrule
\end{tabular}}
\caption{Conflict-type classification on the \conflicts test split ($n=406$), five classes, sorted by accuracy. Every system sees the same inputs; the LLMs receive the paper's prompt verbatim and the decision models a choice question with the same instructions, definitions and examples. Cost: billed cost per 1,000 queries through one gateway (USD). Latency: mean per query. ECE is undefined for text answers. $\Delta$: accuracy difference in points from Gemini~2.5 Flash (thinking off, the paper's setup) and from Gemini~3.8 Flash (thinking low), exact McNemar $p$. $^\S$Its API accepts only a single text field, so it receives the same content as one text (the LLMs' numbered results). $^\P$Free on the gateway at the time of the runs (billed \$0). $^\ddagger$Reported by \citet{cattan2025dragged} on all 458 queries; not comparable query by query.}
\label{tab:main}
\end{table*}

\subsection{Decision models against the LLMs}
\label{sec:main}

Table~\ref{tab:main} gives the main comparison, sorted by accuracy. Asked directly, the \ndm decision models
classify \dmaccmin--\dmaccmax\% of test queries correctly (median \dmaccmedian\%). Gemini~2.5 Flash with the paper's
prompt reaches \gtwofiveacc\% on the same queries. Against it, \ntiedref decision models are statistically tied,
\nbetterref are significantly more accurate (Liquid d1 Ultra, \liquidultravsrefdiff points, $p\liquidultravsrefp$;
Perplexity Decider v1.2, \pplxvsrefdiff, $p\pplxvsrefp$), and \nworseref, ScaleDown Classify-1, is significantly
less accurate (\scaledownvsrefdiff, $p\scaledownvsrefp$). The decision models cost \$\dmcostmin--\$\dmcostmax per
thousand queries (\nfree were free at the time of the runs) against \$\gtwofivecost, and answer in
\dmlatmin--\dmlatmax\,s against \gtwofivelat\,s. Among the \natpar models that match or exceed the paper's LLM,
those that are billed cost \costratiorefmin--\costratiorefmax$\times$ less, and all are
\latratiorefmin--\latratiorefmax$\times$ faster.

The current Gemini~3.8 Flash is stronger: at its lowest thinking level it reaches \gthreeeightacc\%,
\gthreeeightvstwofivediff points above Gemini~2.5 Flash ($p\gthreeeightvstwofivep$), at \costratiorefnow$\times$ its
cost. It is significantly more accurate than \nworsenow of the decision models; the other \ntiednow (Liquid d1
Ultra, Perplexity Decider v1.2 and OpenAI GPT-6 Luna Decisions) trail it by \liquidultravsnowdiffabs to
\lunavsnowdiffabs points without a significant difference. More thinking does not help either Gemini: Gemini~3.8
Flash at its default level scores \gthreeeightdefaultacc\% (\gthinkvsnothinkdiff points against the low level,
$p\gthinkvsnothinkp$) and Gemini~2.5 Flash with thinking \gtwofivethinkacc\% ($p\gtwofivethinkvsnothinkp$).

\paragraph{Reproducing the published number.} On all \nall queries, the scope of the published result, our rerun of
Gemini~2.5 Flash scores \allgtwofive\% against the published 65.3\%. The hosted model may have changed since
publication, and the paper does not specify how results were joined in the prompt or whether thinking was enabled.
We therefore compare every system with our rerun on identical inputs and report the published figure only as
context.

\begin{table}[t]
\centering\footnotesize\setlength{\tabcolsep}{3pt}
\fitwidth{\begin{tabular}{lccccc}
\toprule
F1 & NC & Comp.\ & Opin.\ & Outd.\ & Mis. \\
\midrule
Liquid d1 Ultra & 69.2 & 57.3 & 78.2 & 62.7 & 0.0 \\
Perplexity v1.2 & 67.5 & 59.4 & 76.4 & 67.2 & 11.8 \\
GPT-6 Luna & 70.0 & 67.2 & 63.7 & 72.7 & 0.0 \\
Mammoth-1 & 68.9 & 54.5 & 78.5 & 62.5 & 0.0 \\
Elephant-1 & 69.6 & 59.5 & 70.2 & 58.5 & 0.0 \\
Liquid d1 & 63.0 & 58.4 & 69.8 & 66.2 & 0.0 \\
Jev 1.13 & 64.0 & 62.7 & 72.5 & 65.0 & 10.2 \\
Perplexity Flash v1.2 & 66.9 & 53.6 & 69.1 & 57.7 & 0.0 \\
Jev 1.14 & 59.2 & 63.3 & 73.7 & 61.3 & 9.0 \\
Microsoft D-1 & 51.6 & 58.5 & 77.0 & 59.4 & 4.5 \\
ScaleDown C-1$^\S$ & 60.0 & 42.3 & 57.3 & 27.5 & 0.0 \\
\midrule
Gemini 2.5 (off) & 60.9 & 59.7 & 76.2 & 59.0 & 9.8 \\
Gemini 3.8 (low) & 71.8 & 68.3 & 75.1 & 66.1 & 25.0 \\
\bottomrule
\end{tabular}}
\caption{Per-class F1 on the test split (support: No conflict 143, Complementary 102, Opinions 102, Outdated 55, Misinformation 4): decision models with the flat question, then Gemini~2.5 Flash (thinking off) and Gemini~3.8 Flash (thinking low). $^\S$Its API accepts only a single text field, so it receives the same content as one text (the LLMs' numbered results).}
\label{tab:perclass}
\end{table}

\paragraph{Per class and calibration.} Per class (Table~\ref{tab:perclass}), the models differ most on
\emph{Outdated information}, with F1 from \outdfmin to \outdfmax (Gemini~3.8 Flash: \gthreeeightoutdf).
\emph{Misinformation}, with \nmisitest test instances, defeats every system: no decision model exceeds an F1 of
\misifmax (Gemini~3.8 Flash: \gthreeeightmisif). Calibration varies more than accuracy: the expected calibration
error of the flat question ranges from \eceminimum to \ecemaximum, and \noverconf of the \ndm models are
over-confident on average (mean top probability above accuracy).

\subsection{Flat or tree?}
\label{sec:formulation}

\begin{table}[t]
\centering\footnotesize\setlength{\tabcolsep}{3.5pt}
\fitwidth{\begin{tabular}{lcccc}
\toprule
Decision model & Flat & Tree v1 & Tree v2$^\dagger$ & v2 $-$ flat ($p$) \\
\midrule
Liquid d1 Ultra & 67.0 & 50.2 & 62.8 & $-$4.2 (0.10) \\
Perplexity v1.2 & 66.7 & 56.9 & 69.0 & +2.2 (0.23) \\
GPT-6 Luna & 66.3 & 42.1 & 50.0 & $-$16.3 ({<}0.001) \\
Mammoth-1 & 65.8 & 48.3 & 59.1 & $-$6.7 (0.008) \\
Elephant-1 & 65.3 & 47.0 & 59.4 & $-$5.9 (0.009) \\
Liquid d1 & 62.6 & 44.8 & 58.4 & $-$4.2 (0.13) \\
Jev 1.13 & 61.8 & 47.3 & 63.5 & +1.7 (0.59) \\
Perplexity Flash v1.2 & 61.8 & 46.6 & 59.1 & $-$2.7 (0.29) \\
Jev 1.14 & 59.6 & 45.1 & 62.8 & +3.2 (0.29) \\
Microsoft D-1 & 58.1 & 41.1 & 57.4 & $-$0.7 (0.86) \\
ScaleDown C-1$^\S$ & 50.0 & 37.2 & 39.9 & $-$10.1 (0.001) \\
\bottomrule
\end{tabular}}
\caption{Accuracy (\%) by question format on the test split. The yes/no questions are identical for every model and asked in one request. $^\dagger$Combination order revised on one model's (Jev~1.13) development answers (\S\ref{sec:formulation}). $^\S$Its API accepts only a single text field, so it receives the same content as one text (the LLMs' numbered results).}
\label{tab:formulation}
\end{table}

Decomposing the taxonomy into yes/no questions fails for every model. Tree v1 is significantly less accurate than
the flat question for \nvoneworse of the \ndm models (Table~\ref{tab:formulation}). We had inspected the cause on the
development split with one model, \jev~1.13, before any test run: it answers the \emph{compatible} question
\emph{yes} for opinion topics almost as readily as for complementary ones (mean probability \devcompatopin vs.\
\devcompatcomp), sending \devopintocomp of \devnopin development opinion queries to \emph{Complementary}, while the
\emph{opinion} question separates the two classes well (AUROC \devopinionauroc). On the test split the same model
labels \testopintocomp of the \testnopin \emph{Conflicting opinions} queries as \emph{Complementary} (F1
\treeoneopinf). Its \emph{misinformation} question also shows the yes-bias reported for batched yes/no questions
\citep{deusser2026jev}: the mean probability ranged from \devmisinfolo to \devmisinfohi across all five classes.

Tree v2 asks \emph{opinion} first and treats misinformation as the residual factual conflict. Its order was fitted on
that one model's development answers and is applied unchanged to every model. It is significantly better than the
flat question for none of them, tied for \nvtwotied, and significantly worse for \nvtwoworse. For the model it was
fitted on it is tied with flat (\jevtreeacc\% vs.\ \jevflatacc\%, $p\jevtreevsflatp$) and better calibrated (ECE
\jevtreeece vs.\ \jevece). A repair fitted to one model's errors helped no other model, so we use the flat question
for all of them.

\subsection{Confidence-gated cascades}
\label{sec:cascade}

\begin{table*}[t]
\centering\small\setlength{\tabcolsep}{4.5pt}
\fitwidth{\begin{tabular}{lccccccccc}
\toprule
 & & \multicolumn{4}{c}{25\% escalated} & \multicolumn{3}{c}{50\% escalated} \\
\cmidrule(lr){3-6}\cmidrule(lr){7-9}
Cheap stage (flat) & AUROC & Acc. & Random & $\Delta$ LLM ($p$) & Cost & Acc. & $\Delta$ LLM ($p$) & Cost \\
\midrule
Liquid d1 Ultra & 0.73 & 70.9 & 67.7 & +1.0 (0.63) & 31\% & 70.4 & +0.5 (0.77) & 55\% \\
Perplexity Decider v1.2 & 0.73 & 70.2 & 67.6 & +0.2 (1.00) & 27\% & 70.0 & +0.0 (1.00) & 52\% \\
OpenAI GPT-6 Luna Decisions & 0.65 & 67.0 & 67.2 & $-$3.0 (0.10) & 37\% & 69.0 & $-$1.0 (0.48) & 61\% \\
Baseten Mammoth-1 & 0.70 & 68.7 & 66.8 & $-$1.2 (0.49) & 26\% & 70.2 & +0.2 (1.00) & 50\% \\
Baseten Elephant-1 & 0.72 & 69.2 & 66.4 & $-$0.7 (0.77) & 25\% & 70.0 & +0.0 (1.00) & 50\% \\
Liquid d1 & 0.72 & 67.5 & 64.4 & $-$2.5 (0.14) & 27\% & 69.2 & $-$0.7 (0.55) & 52\% \\
Jev 1.13 & 0.70 & 68.5 & 63.9 & $-$1.5 (0.41) & 30\% & 69.2 & $-$0.7 (0.61) & 54\% \\
Perplexity Decider Flash v1.2 & 0.70 & 67.2 & 63.9 & $-$2.7 (0.13) & 26\% & 68.5 & $-$1.5 (0.26) & 51\% \\
Jev 1.14 & 0.70 & 65.8 & 62.2 & $-$4.2 (0.016) & 30\% & 68.7 & $-$1.2 (0.36) & 54\% \\
Microsoft Decision-1 & 0.76 & 66.0 & 61.1 & $-$3.9 (0.048) & 29\% & 69.2 & $-$0.7 (0.65) & 54\% \\
ScaleDown Classify-1$^\S$ & 0.64 & 56.4 & 55.0 & $-$13.5 ({<}0.001) & 31\% & 62.8 & $-$7.1 ({<}0.001) & 57\% \\
\bottomrule
\end{tabular}}
\caption{Confidence-gated cascade on the test split: the decision model answers every query and the fraction with the lowest top probability goes to Gemini~3.8 Flash (thinking low; 70.0\% alone). AUROC: how well the decision model's confidence separates its correct from its wrong answers. Random: expected accuracy when the same number of queries is escalated at random. $\Delta$ LLM: difference from the LLM alone in points (exact McNemar $p$). Cost: relative to the LLM alone. Thresholds are label-free quantiles, fixed before computing. $^\S$Its API accepts only a single text field, so it receives the same content as one text (the LLMs' numbered results).}
\label{tab:cascade}
\end{table*}

\begin{figure*}[t]
\centering
\includegraphics[width=0.78\textwidth]{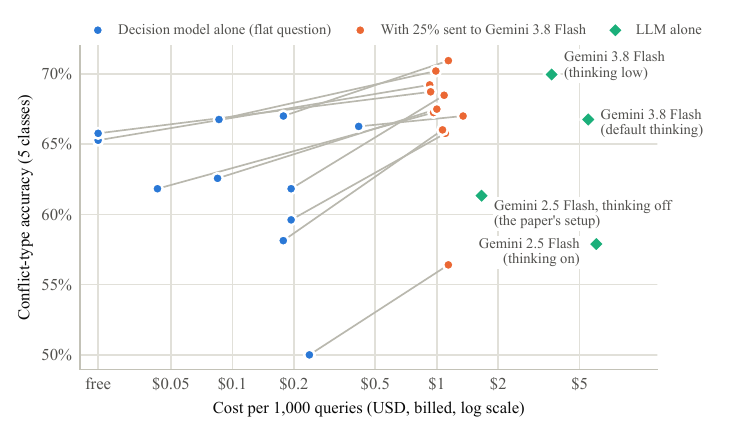}
\caption{Accuracy against cost (log scale) on the test split. Each decision model alone with the flat question
(blue) is joined to the same model with its least confident 25\% of queries sent to Gemini~3.8 Flash (orange).
Diamonds: LLMs alone with the paper's prompt. The two models that were free at the time of the runs are drawn at
the ``free'' tick.}
\label{fig:cascade}
\end{figure*}

Decision models' confidence ranks their own errors, with AUROC from \aurocmin to \aurocmax
(Table~\ref{tab:cascade}), which makes each of them a natural first stage before an LLM. Escalating the least
confident quarter of queries to Gemini~3.8 Flash yields cascades statistically tied with the LLM alone for
\ncasctied of the \ndm models, at \casctiedcostmin--\casctiedcostmax\% of its cost. Escalating by confidence beats
escalating the same number of queries at random for \nbeatrandom of the \ndm models, by up to \cascgainmax points;
the exception is GPT-6 Luna Decisions, whose confidence barely ranks its errors (AUROC \lunaauroc) and whose 25\%
cascade equals random escalation (\casclunatfacc\% vs.\ \casclunatfrand\%). Figure~\ref{fig:cascade} shows every
model alone and as a cascade against the LLMs. A tie at 25\% does not establish equivalence: with \ntest queries, the intervals
still admit losses of a few points relative to the LLM, and showing equivalence at a tight margin would need more
queries than \conflicts provides.

\subsection{Diagnostics}
\label{sec:diagnostics}

We specified three hypotheses and their tests before running them; Tables~\ref{tab:diagnostics}
and~\ref{tab:plant} give the results for every decision model's flat question and for Gemini~3.8 Flash. An identical
rerun sets the noise floor: \ndeterministic of the \ndm decision models return identical answers, and the others
change their predicted type on up to \rerunflipmax\% of queries with nothing changed (Gemini~3.8 Flash:
\diaggmrerunflip\%).

\begin{table*}[t]
\centering\footnotesize\setlength{\tabcolsep}{4pt}
\fitwidth{\begin{tabular}{lccccccc}
\toprule
 & Rerun & \multicolumn{3}{c}{Dates} & \multicolumn{3}{c}{Source order} \\
\cmidrule(lr){3-5}\cmidrule(lr){6-8}
System & Flip & No dates & Recency in code & Outd.\ recall & Flip [95\% CI] & Shuffle 1 & Shuffle 2 \\
\midrule
Liquid d1 Ultra & 5.2 & +1.0 (0.45) & +0.0 (1.00) & 76.4 / 78.2 & 7.4 [5.7, 9.2] & +1.7 (0.25) & +2.2 (0.12) \\
Perplexity Decider v1.2 & 0.0 & +0.5 (0.69) & +0.7 (0.51) & 78.2 / 80.0 & 7.0 [5.3, 8.9] & +0.0 (1.00) & +0.0 (1.00) \\
OpenAI GPT-6 Luna Decisions & 0.0 & $-$0.5 (0.85) & +0.0 (1.00) & 72.7 / 74.5 & 17.9 [15.3, 20.6] & $-$4.7 (0.011) & $-$5.7 (0.006) \\
Baseten Mammoth-1 & 0.0 & $-$0.2 (1.00) & $-$0.2 (1.00) & 72.7 / 70.9 & 5.4 [3.9, 7.0] & +1.2 (0.27) & +2.2 (0.093) \\
Baseten Elephant-1 & 0.0 & +0.5 (0.80) & +0.7 (0.55) & 56.4 / 65.5 & 7.3 [5.5, 9.1] & $-$0.7 (0.68) & $-$1.0 (0.48) \\
Liquid d1 & 9.4 & $-$0.2 (1.00) & $-$0.5 (0.83) & 81.8 / 81.8 & 15.5 [13.1, 18.0] & +0.2 (1.00) & +2.2 (0.25) \\
Jev 1.13 & 2.0 & $-$1.0 (0.34) & +0.2 (1.00) & 72.7 / 74.5 & 11.6 [9.4, 13.8] & +1.5 (0.38) & +0.2 (1.00) \\
Perplexity Decider Flash v1.2 & 0.0 & +0.0 (1.00) & +0.5 (0.79) & 58.2 / 61.8 & 11.7 [9.6, 13.9] & +1.0 (0.60) & +0.2 (1.00) \\
Jev 1.14 & 3.0 & +0.0 (1.00) & +1.0 (0.39) & 76.4 / 76.4 & 10.8 [8.7, 12.9] & +1.0 (0.60) & +0.0 (1.00) \\
Microsoft Decision-1 & 1.5 & $-$0.7 (0.58) & $-$0.2 (1.00) & 69.1 / 67.3 & 17.2 [14.7, 20.0] & +0.2 (1.00) & +0.7 (0.78) \\
ScaleDown Classify-1$^\S$ & 8.4 & $-$2.5 (0.25) & $-$1.5 (0.52) & 20.0 / 25.5 & 42.1 [38.8, 45.4] & $-$3.0 (0.32) & +1.0 (0.78) \\
\midrule
Gemini 3.8 Flash (low) & 3.4 & $-$0.7 (0.51) & $-$0.7 (0.65) & 69.1 / 67.3 & 9.4 [7.5, 11.5] & $-$0.5 (0.85) & $-$2.7 (0.090) \\
\bottomrule
\end{tabular}}
\caption{Diagnostics on the test split, flat question. Each perturbed run is compared with the same system's original run on the same queries. Flip: share of queries whose predicted type changed (\%); for source order, pooled over two shuffles. No dates, Recency in code, Shuffle: accuracy change in points (exact McNemar $p$). Outd.\ recall: recall (\%) on the 55 \emph{Outdated} queries, original / without dates. The identical rerun is the noise floor. $^\S$Its API accepts only a single text field, so it receives the same content as one text (the LLMs' numbered results).}
\label{tab:diagnostics}
\end{table*}

\paragraph{Dates.} One provider documents ordering dates as a known weakness of its model
\citep{typesafe2026jagged}, and we expected date handling to explain errors on \emph{Outdated}. It does not. Removing every date, or adding each result's
recency rank computed in code, changes accuracy significantly in \ndatesig of \ndatetests tests, and recall on
\emph{Outdated} moves little in either direction. The models appear to recognise outdated sources from their
content (phrases such as ``former'' or ``as of'', and conflicting values), not from publication dates.

\paragraph{Source order.} Shuffling the results changes the predicted type on \shuffleflipmin--\shuffleflipmax\% of
queries (pooled over two seeds), above the noise floor for all \ndm models; Gemini~3.8 Flash flips on
\diaggmshuffleflip\%. Paired on the same shuffles, \nflipmore decision models flip significantly more often than the
LLM, \nflipless less often, and \nflipsame no differently. The flips rarely change accuracy: \nshufflesig of
\ndatetests shuffle tests are significant, both for GPT-6 Luna Decisions (\diaglunashuffleonediff and
\diaglunashuffletwodiff points). Order sensitivity is therefore a property of individual models rather than of the
class, and a deployment should test it.

\begin{table}[t]
\centering\footnotesize\setlength{\tabcolsep}{3pt}
\fitwidth{\begin{tabular}{lccccc}
\toprule
System & Hit & F.A. & Youden & Any & AUROC \\
\midrule
Liquid d1 Ultra & 41.7 & 2.1 & +0.396 & 60.4 / 21.9 & 0.85 \\
Perplexity v1.2 & 50.0 & 3.1 & +0.469 & 70.8 / 24.0 & 0.86 \\
GPT-6 Luna & 24.0 & 1.0 & +0.229 & 53.1 / 22.9 & 0.81 \\
Mammoth-1 & 39.6 & 5.2 & +0.344 & 61.5 / 18.8 & 0.81 \\
Elephant-1 & 11.5 & 0.0 & +0.115 & 40.6 / 16.7 & 0.78 \\
Liquid d1 & 30.2 & 5.2 & +0.250 & 64.6 / 35.4 & 0.81 \\
Jev 1.13 & 66.7 & 17.7 & +0.490 & 85.4 / 41.7 & 0.81 \\
Perplexity Flash v1.2 & 25.0 & 1.0 & +0.240 & 57.3 / 25.0 & 0.76 \\
Jev 1.14 & 64.6 & 21.9 & +0.427 & 85.4 / 51.0 & 0.79 \\
Microsoft D-1 & 46.9 & 16.7 & +0.302 & 85.4 / 58.3 & 0.76 \\
ScaleDown C-1$^\S$ & 7.3 & 2.1 & +0.052 & 46.9 / 27.1 & 0.66 \\
\midrule
Gemini 3.8 (low) & 52.1 & 3.1 & +0.490 & 71.9 / 25.0 & -- \\
\bottomrule
\end{tabular}}
\caption{One planted false source in 96 \emph{No conflict} test queries (expected label: \emph{Misinformation}), flat question. Hit: planted copy labelled Misinformation (\%). F.A.: clean copy of the same query labelled Misinformation (\%). Youden: hit $-$ false alarm. Any: planted / clean copy not labelled No conflict (\%). AUROC: P(Misinformation) separating planted from clean copies. $^\S$Its API accepts only a single text field, so it receives the same content as one text (the LLMs' numbered results).}
\label{tab:plant}
\end{table}

\paragraph{Planted false sources.} \conflicts has too few \emph{Misinformation} queries to evaluate. We construct
them: in \nplant \emph{No conflict} test queries whose gold answer appears in some result, we rewrite one such
result so that it states a plausible wrong answer, produced by a third model (GPT-4.1-mini) so that no classifier
judges its own text. The expected label becomes \emph{Misinformation}. As hard labels, the decision models
call \planthitmin--\planthitmax\% of planted copies misinformation (Gemini~3.8 Flash: \plantgmhit\%). Net of false
alarms on the clean copies, their detection (Youden index) ranges from \youdenmin to \youdenmax against
\plantgmyouden for the LLM; paired on the same queries, \nyoudenlower are significantly worse and \nyoudentied are
statistically tied. The probabilities tell a different story: the probability of \emph{Misinformation} separates
planted from clean copies with AUROC \plantaurocmin--\plantaurocmax, at least 0.75 for \nplantaurochigh of the
\ndm. The models notice the planted source but rarely make \emph{Misinformation} their top answer, so a lower
threshold on that probability, chosen on development data, would recover detections.

\paragraph{Version stability.} One provider serves two pinned versions. \jev~1.14 differs from 1.13 by
\versionsdiff points with the flat question ($p\versionsp$) and \versionstreediff with tree v2 ($p\versionstreep$),
with the same failure of tree v1 (Table~\ref{tab:formulation}): the conclusions do not depend on which version is
served.

\subsection{End to end: do cheaper types keep the gain?}
\label{sec:endtoend}

With Gemini~2.5 Flash as the generator, \citet{cattan2025dragged} report that the share of answers following the
expected behaviour rises from 65.7\% without a type (Vanilla) to 75.1\% with types predicted by Gemini~2.5 Flash
(Pipeline) and 88.0\% with gold types (Oracle). Can a decision model be the Pipeline's predictor without losing
this gain?

\paragraph{Setup.} The generator is Gemini~2.5 Flash (thinking off, temperature 0) on the same numbered results.
Vanilla uses their answer prompt verbatim; a type adds its definition and expected behaviour from their paper
(Appendix~\ref{app:prompts}). We generate one answer per test query for Vanilla and for each type (\eenanswers
answers); a predictor receives the answer for the type it predicts, so predictors differ only through their
predictions. As in their study, an LLM judge \citep{zheng2023judging} (Gemini~2.5 Flash) decides whether the answer follows the gold type's
expected behaviour, given the query, the type's description, an adherent and a non-adherent example, but not the
sources; a second prompt checks whether it contains the gold answer (\eenrecall queries). On
\nvalid answers labelled blind by two annotators, who agree on \iaa\% (Cohen's $\kappa=\iaakappa$; \citealp{cohen1960kappa}), the judge agrees with
their majority on \eevalacc\% (\citeauthor{cattan2025dragged}: 89\%; $\kappa=\eevalkappa$, as \humanfollows\% of
answers follow; Appendix~\ref{app:protocol}). Our protocol also specified a second judge, Gemini~3.8 Flash (thinking
low), which agrees less often (\fbvalacc\%); we report both.

\begin{table*}[t]
\centering\small\setlength{\tabcolsep}{4.5pt}
\fitwidth{\begin{tabular}{lcccccccc}
\toprule
 & \multicolumn{2}{c}{Type predictor} & \multicolumn{2}{c}{Judge: Gemini 2.5} & \multicolumn{2}{c}{Judge: Gemini 3.8} & \multicolumn{2}{c}{25\% to Gemini 3.8$^\dagger$} \\
\cmidrule(lr){2-3}\cmidrule(lr){4-5}\cmidrule(lr){6-7}\cmidrule(lr){8-9}
Type given to the generator & Acc. & \$ / 1k & Adh. & $\Delta$ ($p$) & Adh. & $\Delta$ ($p$) & Adh. & $\Delta$ ($p$) \\
\midrule
Vanilla (no type) & -- & -- & 87.9 & $-$3.7 (0.032) & 91.1 & +1.5 (0.47) & & \\
Gemini 2.5 Flash, thinking off$^\ast$ & 61.3 & 1.66 & 91.6 & -- & 89.7 & -- & & \\
\midrule
Gemini 3.8 Flash, thinking low & 70.0 & 3.65 & 92.1 & +0.5 (0.81) & 90.6 & +1.0 (0.52) & & \\
Liquid d1 Ultra & 67.0 & 0.18 & 91.4 & $-$0.2 (1.00) & 88.4 & $-$1.2 (0.42) & 92.1 & +0.5 (0.81) \\
Perplexity v1.2 & 66.7 & 0.09 & 91.4 & $-$0.2 (1.00) & 88.2 & $-$1.5 (0.34) & 92.4 & +0.7 (0.65) \\
GPT-6 Luna & 66.3 & 0.41 & 88.9 & $-$2.7 (0.052) & 89.7 & +0.0 (1.00) & 90.6 & $-$1.0 (0.56) \\
Mammoth-1 & 65.8 & 0$^\P$ & 91.1 & $-$0.5 (0.84) & 88.9 & $-$0.7 (0.70) & 91.4 & $-$0.2 (1.00) \\
Elephant-1 & 65.3 & 0$^\P$ & 90.6 & $-$1.0 (0.54) & 89.9 & +0.2 (1.00) & 91.1 & $-$0.5 (0.83) \\
Liquid d1 & 62.6 & 0.08 & 90.6 & $-$1.0 (0.54) & 89.9 & +0.2 (1.00) & 91.9 & +0.2 (1.00) \\
Jev 1.13 & 61.8 & 0.19 & 91.4 & $-$0.2 (1.00) & 90.1 & +0.5 (0.84) & 92.4 & +0.7 (0.65) \\
Perplexity Flash v1.2 & 61.8 & 0.04 & 90.1 & $-$1.5 (0.24) & 89.9 & +0.2 (1.00) & 90.6 & $-$1.0 (0.50) \\
Jev 1.14 & 59.6 & 0.19 & 92.4 & +0.7 (0.66) & 90.1 & +0.5 (0.84) & 92.4 & +0.7 (0.63) \\
Microsoft D-1 & 58.1 & 0.18 & 91.9 & +0.2 (1.00) & 91.1 & +1.5 (0.29) & 92.4 & +0.7 (0.68) \\
ScaleDown C-1$^\S$ & 50.0 & 0.24 & 87.7 & $-$3.9 (0.011) & 87.9 & $-$1.7 (0.36) & 90.6 & $-$1.0 (0.56) \\
\midrule
Gold type (Oracle) & 100 & -- & 95.3 & +3.7 (0.011) & 96.8 & +7.1 ({<}0.001) & & \\
\bottomrule
\end{tabular}}
\caption{End to end on the test split ($n=406$): Gemini~2.5 Flash answers given the type from the first column (the Pipeline of \citet{cattan2025dragged}). Acc.: type accuracy (\%). \$ / 1k: classification cost per 1,000 queries. Adh.: answers following the gold type's expected behaviour (\%), per LLM judge. $\Delta$: points from the original setup$^\ast$ (exact McNemar $p$). $^\dagger$Cascade of Table~\ref{tab:cascade}, primary judge. $^\S$Single text field (Table~\ref{tab:main}). $^\P$Billed \$0.}
\label{tab:endtoend}
\end{table*}

\paragraph{Results.} With the primary judge the ordering of \citet{cattan2025dragged} reappears
(Table~\ref{tab:endtoend}): Vanilla \eevan\%, Pipeline \eepipe\% (\eepipevsvandiff points, $p\eepipevsvanp$), Oracle
\eeoracle\% (\eeoraclevsvandiff, $p\eeoraclevsvanp$); our judge rates answers higher than theirs, which compresses
the differences. Replacing Gemini~2.5 Flash as the predictor changes little: \eentied of the \ndm decision models are
tied with it (\eedmdiffmin to \eedmdiffmax points), and only the least accurate classifier, ScaleDown Classify-1, is
significantly worse (\eescaledownvspipediff, $p\eescaledownvspipep$). With a quarter of queries escalated to
Gemini~3.8 Flash, \eencasctied of the \ndm cascades are tied with it (\eecascdiffmin to \eecascdiffmax). Gains
concentrate on \emph{Conflicting opinions} (Vanilla \eevanopin\%, Oracle \eeoracleopin\%) and \emph{Outdated}
(\eevanoutd\% to \eeoracleoutd\%); no condition lowers answer recall (\eerecallmin--\eerecallmax\%). With the second
judge, gold types still help (\fboraclevsvandiff, $p\fboraclevsvanp$) but predicted types do not (Pipeline \fbpipe\%,
Vanilla \fbvan\%, $p\fbpipevsvanp$), and \fbntied of the \ndm decision models remain tied with the Gemini~2.5 Flash
Pipeline. The size of the gain from predicted types thus depends on the judge; that a decision model can replace the
LLM as the predictor does not.

\section{Discussion}
\label{sec:discussion}

\paragraph{What decision models offer as conflict classifiers.} Conflict typing sits on the critical path of every
search-augmented answer, so its latency and cost add to every request. Most hosted decision models classify a query
with about ten retrieved pages in under half a second, for at most \$\dmcostmax per thousand queries, with the accuracy of the LLM used in the original study. When the accuracy of a
current LLM is required, a decision model's confidence is usually good enough to route only its uncertain queries to
the LLM, which recovers most of the gap at a quarter to a third of the LLM's cost. End to end, the cheaper types cost
no measurable answer quality (\S\ref{sec:endtoend}). 

\paragraph{Choosing among decision models.} The class is not uniform. Accuracy spans \dmaccmin--\dmaccmax\%, and the
model at the bottom is significantly worse than the paper's LLM. Two properties that matter in deployment vary
independently of accuracy: calibration (ECE \eceminimum--\ecemaximum) and how well confidence ranks errors, which
decides whether a cascade helps at all. One of the most accurate models has confidence that is no better than random for
escalation. A model for this task should therefore be chosen on all three, measured on the deployment's own data,
not on accuracy alone.

\paragraph{How to ask.} The direct choice question was the reliable format for every model. The yes/no
decomposition looked natural but failed for all of them; the repaired tree encodes choices fitted to one model on a
small development split and did not help any other. For a new decision model we would start from the flat
question.

\paragraph{What the diagnostics change.} Dates turned out not to matter: outdated sources are recognised
from their content. Sensitivity to source order is real but differs so much between models that it has to be
measured for each one. A single false source is the subtle case, visible in the probabilities but rarely in the
top label. With
\nmisitest natural \emph{Misinformation} test queries, that type is the one a deployment would most need to
calibrate on its own data.

\paragraph{Limitations.}
(i) \conflicts is small: with \ntest test queries, differences under about five points are not significant, and the
cascades can be shown close to, but not formally equivalent to, the LLM. (ii) The label distribution reflects how the
queries were selected, not real search traffic, and the dataset has only five natural \emph{Misinformation}
queries; our planted variant tests one kind of misinformation (an entity or value substituted in a single source).
(iii) The census covers the decision models one gateway served on one day. Hosted models change without notice:
several identifiers are unversioned, two models were free during our runs, and our rerun of Gemini~2.5 Flash scores
below the published figure. Providers
do not disclose model sizes or training data, so we cannot explain the differences between models. (iv) The end-to-end test uses one generator and LLM judges, validated on \nvalid answers, that do not see the
search results; as in the original study, the primary judge is the generator's own model, which may favour its
answers \citep{panickssery2024llm}. The size of the gain from predicted types depends on the judge, and we do not
measure factual grounding, the third metric of \citet{cattan2025dragged}. (v) Not every system is deterministic: identical requests changed
up to \rerunflipmax\% of a decision model's predictions, and \diaggmrerunflip\% of Gemini~3.8 Flash's.

\section{Conclusion}
\label{sec:conclusion}

We tested every typed decision model hosted on one gateway as a classifier of knowledge-conflict types, using
the benchmark authors' protocol and the same inputs as the LLMs. Most match the LLM of the original study. They do
so at a small fraction of its cost and latency. They trail the
current LLM, but for most of them a confidence-gated cascade recovers that gap at a quarter to a third of the LLM's
cost, and given to a generator, their types yield answers that follow the expected behaviour as often as the LLM's.
 Future work should measure the gain from predicted types with judges that see the
sources, and test whether a decision model's probabilities can steer a generator between expected behaviours when
the type is uncertain.

\section*{Ethics and Broader Impact}

The planted-misinformation variant rewrites one retrieved page per query to state a wrong answer. It uses public data and factual substitutions on
non-sensitive questions, and is released with clear labels. All queries in \conflicts are safe by construction \citep{cattan2025dragged}. Cheap conflict
classification could be misused to suppress minority viewpoints; the expected behaviour for \emph{Conflicting
opinions} is to present all of them.

\bibliography{refs}
\bibliographystyle{acl_natbib}

\appendix
\section{Prompts and Questions}
\label{app:prompts}

\paragraph{LLM prompt.} The conflict-type prompt of \citet{cattan2025dragged}, copied from the paper's source with
only the \LaTeX{} bold markers removed. The asterisks are the prompt's own Markdown emphasis and were sent
as written. \texttt{\{search\_results\}} is filled with the numbered results.

\begin{lstlisting}
You are tasked with analyzing the search results provided for a given query and classifying the type of knowledge conflict present (if any).
Consider the query and the search results carefully, then classify the conflict into *one* of the following categories, using the descriptions and examples provided below.

1. No Conflict: The search results refer to the same concept and are in agreement. Differences are superficial, such as variations in detail or granularity.
    *   *Example:*
        *   *Query:* What is the meaning of the name Apoorva?
        *   *Search Results:* Unique, Quite new, Not seen before
2.  Complementary Information: The question is underspecified or allows for multiple valid perspectives or answers that do not contradict each other. All provided answers can be considered correct.
    *   *Example:*
        *   *Query:* Is public transportation faster than driving in cities?
        *   *Search Results:* Depends on the city and situation (e.g., rush hour vs. off-peak), specific routes matter, availability of parking is relevant.
3.  Conflicting Opinions or Research Outcomes: The query addresses a subjective or contentious topic, leading to genuinely opposing viewpoints or contradictory research findings.
    *   *Example:*
        *   *Query:* Is online learning as effective as traditional classroom learning?
        *   *Search Results:* Some sources argue "yes," citing flexibility and accessibility. Others argue "no," emphasizing the importance of in-person interaction.
4.  Conflict Due to Outdated Information: The question has a verifiable factual answer, but the search results contain conflicting information due to changes over time.
    *   *Example:*
        *   *Query:* Do Tesla and X Corp. have the same CEO?
        *   *Search Results:* Some articles say yes (dated before the change), others say no (dated after the change).
5.  Conflict Due to Misinformation: The question has a verifiable factual answer, but some of the search results contain factually incorrect or misleading information.
    *   *Example:*
        *   *Query:* What is the capital of Israel?
        *   *Search Results:* One source correctly states Jerusalem, while another incorrectly states Tel Aviv.


Input:
*   QUERY: {query}
*   SEARCH RESULTS: {search_results}

Output:
Please provide your response as a JSON object with the following fields:
*   explanation: (String) A brief explanation of why you chose this category.
*   category: (Integer) The number (1-5) corresponding to the category of the knowledge conflict.
\end{lstlisting}

\paragraph{Decision models, flat.} One choice question. Its instruction is the first paragraph of the prompt above; its options
are the five category definitions with their examples, word for word.

\begin{lstlisting}
type: choice
instructions: You are tasked with analyzing the search results provided for a given query and classifying the type of knowledge conflict present (if any).
Consider the query and the search results carefully, then classify the conflict into *one* of the following categories, using the descriptions and examples provided below.
options: the five category definitions of the prompt above, each with its example, keyed no_conflict, complementary, conflicting_opinions, outdated, misinformation
\end{lstlisting}

\paragraph{Decision models, tree.} Five yes/no questions asked in one request over the same state.

\begin{lstlisting}
differ (yes/no): Beyond superficial differences in wording, detail or granularity, do the search results give different answers to the query?
compatible (yes/no): Could all the different answers in the search results be correct at the same time, for example because the query is underspecified or allows several valid answers?
opinion (yes/no): Is the query a subjective or contentious topic on which the search results present opposing viewpoints or contradictory research findings?
outdated (yes/no): Does the query have a verifiable factual answer on which the search results disagree because the answer changed over time, so some results are outdated?
misinformation (yes/no): Does the query have a verifiable factual answer on which some of the search results are factually incorrect or misleading?
\end{lstlisting}

\paragraph{Planted wrong answers.} Prompt sent to GPT-4.1-mini (temperature 0) to obtain a plausible wrong answer
for each \emph{No conflict} query whose gold answer appears in a result. The answer is then replaced, as whole words
and case-insensitively, in the title, snippet and text of one such result chosen with a fixed seed. Queries whose
gold answer is \emph{yes} or \emph{no} are excluded.

\begin{lstlisting}
Question: {question}
Correct answer: {answer}

Give one plausible but incorrect answer to the question, of the same type and format as the correct answer (a different name, place, number or date). Reply with the incorrect answer only.
\end{lstlisting}

\paragraph{End to end: generation.} Vanilla is the answer prompt of \citet{cattan2025dragged} (their Figure~3)
verbatim. For a type, the line below is inserted before the query; the description is the type's definition and
expected behaviour from their \S2, with citations and worked examples removed.

\begin{lstlisting}
Write a high-quality and concise answer for the given question using only the provided search results.
For each sentence, cite the corresponding sources using [1][2][3].

TYPE OF KNOWLEDGE CONFLICT IN THE SEARCH RESULTS: {description}

QUERY: {query}
SEARCH RESULTS: {search_results}
\end{lstlisting}

\paragraph{End to end: judges.} The behaviour judge, shown for \emph{Misinformation}; for the other types the
description and the two examples change. The examples are those of \citet{cattan2025dragged}, Table~3, except for
\emph{No conflict}, which has none there and for which we wrote one pair. The answer-recall judge follows.

\begin{lstlisting}
You are evaluating whether a response to a search query follows the expected behavior for the type of knowledge conflict among the query's search results.

CONFLICT TYPE: Conflict due to misinformation. The retrieved documents provide answers that are not mutually compatible, where at least one source contains information that is likely false, misleading, or inaccurate. The expected behavior for this category is to disregard inaccurate sources and to provide a response grounded in reliable and verified information.

EXAMPLES:

Query: When did season 5 of prison break come out?
Response: Season 5 of Prison Break premiered on April 4, 2017, on Fox [3]. The season was released on May 30, 2017 [4].
Verdict: does not follow the expected behavior. Reason: The response tries to reconcile the sources, although source [4] is incorrect.

Query: When did season 5 of prison break come out?
Response: Season 5 of Prison Break premiered on April 4, 2017 [2][3]. The first episode aired on this date [2].
Verdict: follows the expected behavior. Reason: The response provides a clear and coherent answer to the question.

RESPONSE TO EVALUATE:
Query: {query}
Response: {response}

Does the response follow the expected behavior for this conflict type? Judge how the response handles the sources, as the expected behavior describes; do not judge its length or writing quality.
Reply with a JSON object: {"explanation": "<one or two sentences>", "adheres": true or false}
\end{lstlisting}

\begin{lstlisting}
You are checking whether a response contains the correct answer to a question.

QUESTION: {query}
CORRECT ANSWER: {gold answer}
RESPONSE: {response}

Does the response state the correct answer? Count paraphrases, equivalent formats and answers with more detail as correct; the response may also mention other information.
Reply with a JSON object: {"explanation": "<one sentence>", "includes_answer": true or false}
\end{lstlisting}

\section{Protocol Record}
\label{app:protocol}

The development/test split, the prompt and question texts and the combination rules were frozen with recorded file
hashes before any test-set call. Changes after that point were limited to the following, each logged with its reason
before results were computed: (1) Gemini~2.5 Flash with thinking added as a sensitivity check; (2) the answer parser
made lenient for category statements outside the requested JSON, applied to stored answers without new calls (no
strictly formatted answer changed); (3) Gemini~3.8 Flash added at two thinking levels; (4) the cascade, its
operating points and its controls; (5) the diagnostics, with one amendment before any planted run: whole-word
matching and the exclusion of yes/no answers, after substring matching was found to alter unrelated numbers;
(6) a second \jev version and four further decision models, run with the frozen questions; (7) the remaining
five decision models listed by the gateway, completing the census, with the diagnostics extended to every decision
model's flat question, and one amendment before any run: Classify-1 receives the same content as a single text
field, which its API requires; (8) the end-to-end test, specified before any generation call: generator, prompts,
judges, the \nvalid-answer validation sample and its threshold, and a second judge (Gemini~3.8 Flash) to be reported
with the first. Prompts were checked on the development split for errors only; none was changed after scoring. All
runs are reported.

\paragraph{Judge validation.} The \nvalid answers were drawn with a fixed seed, 20 per gold type and about half from
Vanilla, and shown to the annotators with the query, the gold type's description and expected behaviour, and the
answer, without the judge's verdict or the condition. Annotator 2's sheet showed another item's answer for one item;
that label is excluded. The primary judge agrees with the annotators' majority (a tie counting as \emph{follows},
as fixed before labelling) on \eevalacc\%, with annotator 1 on \eevalaccone\% and with annotator 2 on
\eevalacctwo\%; the second judge on \fbvalacc\%, \fbvalaccone\% and \fbvalacctwo\%. Code,
splits, prompts, cached responses and per-query outputs are released.

\end{document}